\documentclass[conference]{IEEEtran}
\IEEEoverridecommandlockouts
\usepackage{cite}
\usepackage{algorithmic}
\usepackage{graphicx}
\usepackage{makecell}
\usepackage{sidecap}
\usepackage{textcomp}
\usepackage{multirow}
\usepackage{float} 
\usepackage[numbers,sort&compress]{natbib} 
\usepackage{xcolor}
\usepackage{amsmath,amssymb,amsfonts}
\usepackage{algorithmic}
\usepackage{graphicx}
\usepackage{textcomp}
\usepackage{xcolor}
\def\BibTeX{{\rm B\kern-.05em{\sc i\kern-.025em b}\kern-.08em
    T\kern-.1667em\lower.7ex\hbox{E}\kern-.125emX}}

\begin{document}
\title{ProDehaze: Prompting Diffusion Models Toward Faithful Image Dehazing}

\author{
\IEEEauthorblockN{1\textsuperscript{st} Tianwen Zhou*}
\IEEEauthorblockA{\textit{Department of Computer Science} \\
\textit{University College London}\\
London, United Kingdom \\
tianwenzhou0521@ieee.org}
\and 
\IEEEauthorblockN{2\textsuperscript{nd} Jing Wang}
\IEEEauthorblockA{\textit{R\&D Center Beijing Lab} \\
\textit{Sony China Ltd.}\\
Beijing, China \\
JingD.Wang@sony.com}
\and 
\IEEEauthorblockN{3\textsuperscript{rd} Songtao Wu}
\IEEEauthorblockA{\textit{R\&D Center Beijing Lab} \\
\textit{Sony China Ltd.}\\
Beijing, China \\
Songtao.Wu@sony.com}
\and 
\IEEEauthorblockN{4\textsuperscript{th} Kuanhong Xu}
\IEEEauthorblockA{\textit{R\&D Center Beijing Lab} \\
\textit{Sony China Ltd.}\\
Beijing, China \\
Kuanhong.Xu@sony.com}
}

\maketitle

\begingroup
\renewcommand\thefootnote{}\footnote{\textbf{*} This work was conducted during an internship at Sony China Ltd.}
\addtocounter{footnote}{-1}
\endgroup

\begin{abstract}

Recent approaches using large-scale pretrained diffusion models for image dehazing improve perceptual quality but often suffer from hallucination issues, producing unfaithful dehazed image to the original one. To mitigate this, we propose ProDehaze, a framework that employs internal image priors to direct external priors encoded in pretrained models. We introduce two types of \textit{selective} internal priors that prompt the model to concentrate on critical image areas: a Structure-Prompted Restorer in the latent space that emphasizes structure-rich regions, and a Haze-Aware Self-Correcting Refiner in the decoding process to align distributions between clearer input regions and the output. Extensive experiments on real-world datasets demonstrate that ProDehaze achieves high-fidelity results in image dehazing, particularly in reducing color shifts. Our code
is at https://github.com/TianwenZhou/ProDehaze.

% Image dehazing is a challenging and ill-posed problem in image restoration. In this paper, we propose Prompting Diffusion Models Toward Faithful Image Dehazing (ProDehaze), a novel framework that leverages selective internal priors to guide external generative priors, enabling more faithful and controlled dehazing while minimizing hallucination artifacts. Specifically, we introduce a Structure-Prompted Restorer, which utilizes internal structural priors from the latent space to selectively emphasize regions with structural information. Furthermore, we propose a Haze-Aware Prompted Self-Correction Refiner, designed to align distributions between clearer regions of the input and the reconstructed output, effectively mitigating color shifts and deprioritizing haze-dominated areas. Extensive experimental evaluations demonstrate that ProDehaze significantly outperforms state-of-the-art methods on multiple real-world test sets. Ablation studies further validate the effectiveness of each proposed module.

\end{abstract}

\begin{IEEEkeywords}
Image Restoration, Diffusion Models, Image Dehazing, Visual Prompt Learning
\end{IEEEkeywords}
\vspace{-0.3cm}
\section{Introduction}
\vspace{-0.1cm}
Restoring a clean image from a hazy input is a long-standing yet challenging ill-posed problem. Dehazing is crucial for various applications, including autonomous driving \cite{b39}, where clear visibility is essential for safe navigation, and aerial photography \cite{b40} or remote sensing \cite{b41}, which require haze removal for accurate environmental monitoring. Significant progress has been made through the advent of deep learning-based methods \cite{b14,b15,b17,b18,b8}, which employs deep neural networks to either estimate parameters of the Atmospheric Scattering Model (ASM) \cite{b35,b36} or directly restore the dehazed image. These methods rely on synthetic hazy-clean image pairs for training, 
which often generalizes poorly to real scenes, as synthetic haze may not adequately represent the complexity and diversity of actual conditions. 
Recently, diffusion models \cite{b20,b21} have shown promising results in various image restoration tasks \cite{b23,b24,b25}, including dehazing \cite{b7, b22}.

Several studies \cite{b26,b27,b25} have successfully utilized \textit{external} priors—specifically, generative priors embedded in large-scale pretrained diffusion models 
—to enhance the perceptual quality of restored images significantly. Unlike traditional learning-based methods, 
pretrained diffusion models exploit the inherent distribution of natural images, leading to superior generalization and improved photo-realism. Notably, powerful text-to-image (T2I) models, such as Latent Diffusion Models (LDMs) \cite{b21}, are often used to provide these external priors. The dehazing process is learned in the latent space through finetuning a pretrained denoising UNet. However, despite achieving high-quality results, these methods remain susceptible to \textit{hallucination} issues \cite{b37}. Restored image may lack faithfulness, with inconsistencies in content or details compared to the original images, especially for severely degraded inputs. To address this, some methods \cite{b5} finetune the decoding process of the autoencoder in LDMs, yet they still struggle with color shifts and detail inconsistencies.

To mitigate hallucination issues in real-scene dehazing, this paper explores the use of \textit{internal} image priors—statistics derived from within an input image—to direct the \textit{external} priors of LDMs for failthful dehazing. 
We propose that these internal priors should be \textit{selective} enough to identify critical areas in the input image essential for faithful dehazing. By prompting these priors into the pretrained LDMs, we aim to shift the models' knowledge toward these areas, thereby enabling reliable restoration of hazy images.

Specifically, in the semantically-rich \textbf{latent space} of LDMs \cite{b21}, we propose spatial structure signals as internal priors to guide the model's emphasis on structure-rich areas. It is unveiled that these areas are more decomposable in the frequency domain than in the spatial domain \cite{b78}. Therefore, we derive prompts through frequency decomposition to indicate spatial cues for key areas. The prompts are then injected into the proposed Structure-Prompted Restorer (SPR) to enable the model to capture spatial correlations and enhance fidelity of the corresponding details.
In the \textbf{decoding process}, we introduce a novel Haze-Aware Self-Correcting Refiner to ensure faithful decoding. We propose using a haze-aware prior to enhance the representation of clearer areas while minimizing the impacts of haze-dominant areas. A self-correction mechanism is then employed to promote alignment between the clearer regions of the input hazy image and the output dehazed image. This is achieved by inserting self-attention blocks into the decoder and re-weighting the attention maps using the haze-aware priors. Extensive experiments demonstrate the effectiveness of this self-correcting mechanism in reducing color shifts.

The contributions of this paper can be summarized as:
\begin{itemize}
    \item We introduce using internal image priors to guide the external knowledge of pretrained LDMs toward critical areas in hazy images, aiming to mitigate hallucination issues and thereby, produce failthful dehazed images.
    % \item We propose \textbf{PRIDE} (Prompting Latent Diffusion Models with Selective Internal Priors for Image Dehazing), a novel framework that effectively utilize selective internal priors to guide external generative priors to achieve high-quality and faithful dehazing results. 

    \item In the latent space, we prompt structural priors into the pretrained diffusion model to improve its capture of spatial correlations towards structrual-rich areas, thereby reliably restoring the details of these areas.
    % \item We leverage high-frequency components derived from Haar Discrete Wavelet Transformation (DWT) as structural internal priors in the VAE latent space. This strategy allows PRIDE to prioritize regions with detailed structures that remain consistent across RGB and latent spaces, while being less affected by haze.

    \item In the decoding process, we employ haze-aware priors and a self-correction mechanism to align distributions between clearer input regions and output image, effectively reducing color shifts.
    % \item To address the randomness inherent in the sampling process of latent diffusion models, we introduce a Haze-Aware Self-Correcting Refiner. This module incorporates an internal haze-aware prior to deprioritize haze-dominant regions and refine features in the decoder space.

\end{itemize}
\section{Related Works}
% \textcolor{red}{to be reduced! List/summarize RELATED WORKS! Don't list so many UNRELEVANT works!}
\subsection{Image Dehazing} 
%\textcolor{red}{(No more than 2 or 3 sentences to summarize the prior-based methods!)}
%Prior-based dehazing methods utilize hand-crafted priors, including the Dark Channel Prior (DCP) \cite{b33}, Non-Local Prior \cite{b58}, and Color Attenuation Prior \cite{b57}, to estimate key parameters like transmission maps and depth. While these approaches improve dehazing performance, they struggle to generalize to real-world scenarios due to the complexity of haze distributions.

%\textcolor{red}{(Introduce deep-learning based methods in one paragraph! Don't categorize into end-to-end and non end-to-end! Mimic the writing of RIDCP or FCB. Include typical/baseline works (using synthetic pairs, ASM model) and real-scene works!)} 
While prior-based dehazing methods utilize hand-crafted priors \cite{b33}, deep learning approaches restore haze-free images in a data-driven manner. Early works employed neural networks to estimate intermediate parameters \cite{b61, b60} of ASM or directly predict dehazed images \cite{b14, b15}. However, these methods often degrade in performance with real-world hazy images.
%, as seen in the approaches of Zhang \textit{et al.} \cite{b61} and Ren \textit{et al.} \cite{b60}. 
%Alternatively, fully end-to-end networks, such as AODNet
%by Li \textit{et al.} 
%\cite{b14} and DehazeFormer \cite{b15}, directly predict dehazed images without relying on intermediate representations. However, these data-driven approaches suffer from performance degradation when applied to real-world hazy images. 
%Recent methods have introduced advanced architectures like multi-scale perceptual networks \cite{b62, b63}, frequency-based techniques \cite{b9}, and Transformer-based models \cite{b15, b11} to enhance dehazing performance. 
Recently, deep generative models, including CycleGAN-based domain adaptation strategies \cite{b73},
%like GANs \cite{b72} 
have been explored for real-scene dehazing.
%such as generating realistic hazy data tailored to real haze domains \cite{b71} or leveraging CycleGAN-based domain adaptation strategies \cite{b73}.
%, but these strategies often suffer from artifacts that hinder training effectiveness. 
% Other real-scene methods integrate knowledge specific to real-world conditions through loss functions or tailored network architectures. For example, some approaches adopt prior-based loss training \cite{b74}, while others utilize high-quality codebooks pretrained on real datasets to adapt to real-scene complexities \cite{b12}. However, these methods still struggle with domain generalization, which hampers their robustness and applicability in diverse real-world dehazing scenarios.
Wu \textit{et al.} \cite{b12} further utilize high-quality codebook priors from large-scale datasets to handle real-scene complexities. Nonetheless, these methods still face challenges with domain gap, hindering their real-world applicability.

%\textcolor{red}{(Don't elaborate! Briefly use 2 sentences to summarize! Move the diffusion methods into sec.B)} 
%Despite these advancements, most deep learning methods heavily rely on synthetic hazy-clean image pairs, leading to domain-specific solutions that generalize poorly to real-world scenarios. 

\subsection{Diffusion Models for Image Restoration} 
%\textcolor{red}{Introduce diffuison, especially latent diffusion dehazing which utilize 'pretrained knowledge'. If there are not many dehazing works, you can firstly introduce restoration works, next dehazing. Dont't focus on 'conditioning' (condition isn't relevant to your works, it's relevant to FCB)!} 

Diffusion models \cite{b20} have emerged with impressive results in image restoration tasks \cite{b23,b26,b66,b27}, including dehazing \cite{b7,b22}.
%, due to their ability to generate high-quality perceptual outputs. 
Early works \cite{b7} trained these models from scratch, limiting their ability to leverage the power of pretrained generative priors. Recent research has utilized pretrained LDMs \cite{b21} for restoration. For instance, Lin \textit{et al.} \cite{b27} introduced DiffBIR, which integrates degradation removal and information regeneration modules within pretrained LDMs. Wang \textit{et al.} \cite{b26} advanced this with a Controllable Feature Wrapping (CFW) module to balance fidelity and perceptual quality.
%of the reconstruction result, while Yang \textit{et al.} \cite{b25} proposed a pixel-aware cross-attention mechanism to better capture local structures. 
Yang \textit{et al.} \cite{b75} explored LDMs for dehazing, by investigating the semantic properties of hazy images in the latent space.
However, these methods rely on generative priors obtained from external databases without effectively incorporating visual cues from input images, which can lead to hallucination artifacts and reduced fidelity in real-world scenarios.

%Despite these advancements, existing diffusion-based restoration methods primarily rely on external generative priors without incorporating carefully selected guidance from the input image. This limitation can lead to hallucination artifacts and compromise the fidelity of restored outputs, particularly in complex real-world scenarios.

\subsection{Visual Prompt Learning}
Prompt learning, originally developed in natural language processing (NLP) \cite{b67}, has recently been adapted to low-level vision tasks to leverage task-specific priors \cite{b2,b3,b76,b77,b69,b70}. Methods like SelfPromer \cite{b69} and PromptRestorer \cite{b70} utilize image priors, such as depth consistency or degradation-specific features, to guide restoration. However, these priors often lack selectivity in identifying critical areas in the input image for pretrained models to focus on, potentially diluting their effectiveness. Our work addresses this issue by prompting LDMs with highly selective image priors, enabling more faithful dehazing.
%these methods often incorporate internal priors without selectivity, potentially diluting their effectiveness. To address this limitation, our framework integrates selective internal priors with pretrained generative priors from LDMs, enabling more reliable and faithful dehazing.

%\textcolor{red}{(Make it succint! Don't elaborate! You work is novel in 'selective', not 'internal' (depthmap is also internal)!)} 

\section{The proposed Method}

\subsection{Overview}
Our proposed prompts selectively direct the external knowledge of pre-trained LDMs to image regions critical for faithful dehazing. We designed two types of prompts: 
%we have designed two types of prompts that extract selective internal signals from the input image, which are then injected into the latent space and decoder of the LDMs. 
in the latent space, the regions with fine structures are prioritized through a Structure-prompted Restorer to ensure faithfulness (Sec \ref{LDM}); during subsequent decoding, heavily degraded areas are deprioritized using a Haze-aware Self-correcting Refiner (Sec \ref{HCR}).
%Our proposed method aims to harnesses external priors captured in pre-trained LDMs and investigates effective selective internal priors for single image dehazing. 
An overview of the framework is presented in Fig. \ref{overview}. 
%As shown in the figure, beyond the pretrained LDM—which comprises the Latent Denoising UNet and VAE Encoder—our approach incorporates two key modules: a \textbf{S}emantic-aware \textbf{F}eature \textbf{I}njection module (SFI) , as described in Section III.B and a Self-\textbf{C}orrection Prompted \textbf{H}aze-aware \textbf{R}efine \textbf{N}etwork (\textbf{CHRN}), as described in Section III.C.

% \begin{figure}[t]
%    \centerline{\includegraphics[width=9cm]{images/dwt.png}}
%    \caption{Visualization of the RGB image and its frequency-modulated counterpart, \(x_{\text{high}}\), for (a) the ground-truth (GT) and (b) the hazy image. The difference maps (c) represents the absolute difference between GT and Hazy, revealing that \(x_{\text{high}}\) exhibits significantly smaller variations between haze-affected and clean image than the RGB, highlighting the robustness of our prior. Additionally, (d) illustrates the feature differences extracted by the adapter through activation maps, comparing cases with and without the proposed SPR. The results demonstrate that SPR effectively enhances the structural information of the input image, emphasizing its structures, contours, and edges.
% }\label{fig:dwt}
% \end{figure}

\begin{figure*}[htbp]
    \centering

        \includegraphics[width=\textwidth]{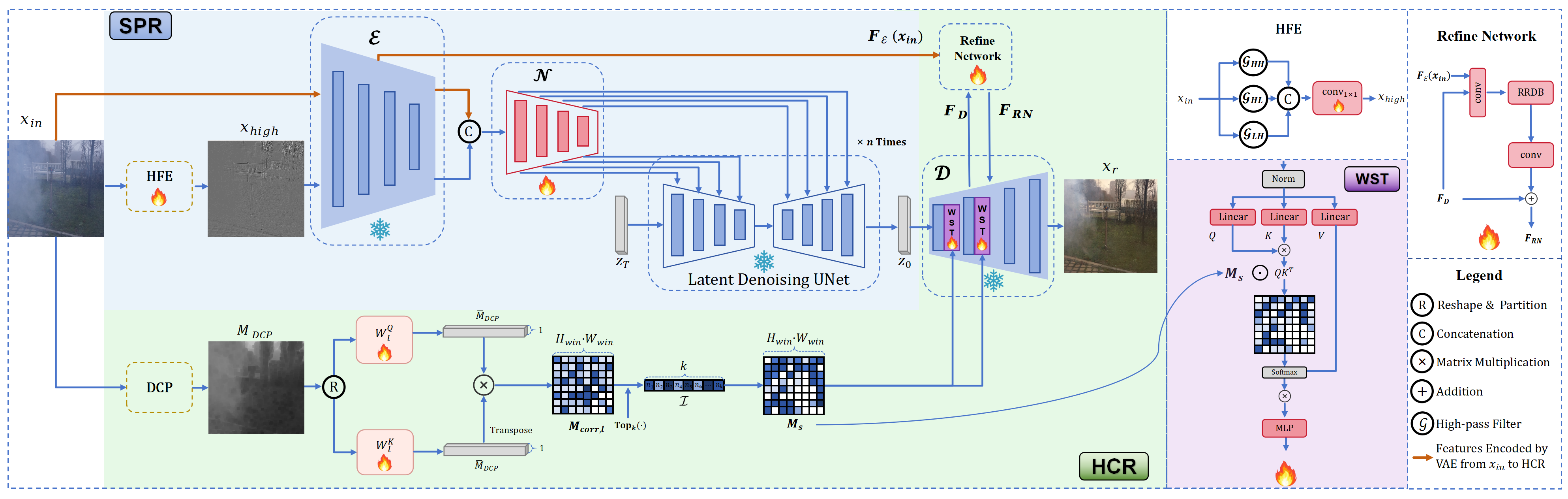}\label{overview}
        \vspace{-0.3cm}  

%\textcolor{red}{(Re-write the fig title. Don't repeat what is already written in Methodology. The purpose is to help the reader walk through your figure and therefore, try to make the explaination more logical and more high-level. For example, explain the framework is finetuned in two phase. In 1st phase, we train SPR in the latent space by structural prompt (then briefly explain the prompt is generated by dwt module and concat with input to inject into an adapter...). Next, we finetune the HCR in the decoding process by haze-aware prompt (then briefly explain how the pompt is generated and injected into WST, and how to train WST and refiner together))} 
\vspace{-0.5cm}
\caption{\textbf{Framework of ProDehaze.} We employ a two-phase finetuning strategy for faithful dehazing. In the first phase, we train the Structure-Prompted Restorer (SPR) in the latent space using a structural prompt generated by a Haar Feature Extractor (HFE) from the hazy input \(x_{in}\). It is then concatenated with the latent representation of \(x_{in}\) and injected into the trainable adapter $\mathcal{N}$ to provide structural guidance. In the second phase, we finetune the Haze-aware Self-Correcting Refiner (HCR) in the decoding process. The haze-aware prompt, initialized by Dark Channel Prior (DCP), produces a sparse mask $M_s$ that emphasizes the clearer areas in \(x_{in}\). It is used to modulate the attention map of the window swin transformer (WST) in the decoder $\mathcal{D}$. Finally, $\mathcal{D}$ and the refine network are jointly trained for better alignment between the clearer regions in \(x_{in}\) and the output \(x_{r}\).}
\vspace{-0.5cm}
    \label{overview}
\end{figure*} 

\subsection{Structure-prompted Restorer (SPR)}\label{LDM}
In the latent space where pre-trained diffusion models encode rich semantics \cite{b21}, we identify \textit{selective} internal priors—fine structures, contours, and edges—that provide essential visual cues for semantic meaning. These elements, as shown in Supplementary Materials II, are more distinct in frequency domain than in image space. Moreover, high-frequency structures are more resilient to haze, as haze primarily carries low-frequency information \cite{b7}. Therefore, we propose using High-frequency prompts to guide the dehazing process.
%To emphasize these areas and simultaneously suppress the negative effects of haze, which primarily resides in low-frequency bands, we propose using high-frequency prompts to guide the dehazing process. 

%Unlike previous works \cite{b7,b25}, our prompt is designed to teach the diffusion model where to attend to.  To achieve this, we directly apply the Haar Discrete Wavelet Transform (DWT) to the input hazy image \( x_{in} \) to obtain its frequency-modulated counterpart \( x_{high} \).

%Previous works \cite{b31} have shown that the self-attention layers of diffusion models effectively capture spatial correlations in images. To unlock this capability, we directly extract high-frequency prior from the input hazy image \( x_{in} \) and inject it, concatenated with \( x_{in} \), into the denoising UNet. 
%Specifically, we apply the Haar Discrete Wavelet Transform (Haar DWT) to the input hazy image \( x_{in} \) to obtain its frequency-modulated counterpart \( x_{high} \). Three high-pass filters  $\mathcal{G}^{LH}$, $\mathcal{G}^{HH}$, and $\mathcal{G}^{HL}$ \cite{b30} of size $2 \times 2$ are selected to extract high frequency components and  $x_{high}$:
Specifically, we utilize the Haar Feature Extractor (HFE), which is the Haar Discrete Wavelet Transform (Haar DWT) \cite{b80} followed by a learnable convolution kernel. The Haar DWT extracts three high-frequency components (\(\mathcal{G}_{LH} * x_{in}\), \(\mathcal{G}_{HH} * x_{in}\), and \(\mathcal{G}_{HL} * x_{in}\)) from the input \(x_{in}\). These components are then fused and processed with a point-wise convolution kernel to produce the high-frequency feature:
\begin{equation}
   x_{high} = \text{conv}_{1\times 1}[(\mathcal{G}_{LH} * x_{in})\oplus (\mathcal{G}_{HH} * x_{in})\oplus (\mathcal{G}_{HL} * x_{in})]
   \label{eqa:kernel}
\end{equation}
Here, \( * \) denotes a convolution operation, \(\oplus\) indicates channel-wise concatenation, and \(\text{conv}_{1\times 1}\) represents the point-wise convolution kernel. To enable the pretrained denoising UNet to capture high-frequency structural signals, we concatenate \( x_{high} \) and \( x_{in} \) in the latent space of LDM: 
\begin{equation}
    c_{f} = \mathcal{E}(x_{in}) \oplus \mathcal{E}(x_{high})
    % c_{f} = [\mathcal{E}(x_{in}), \mathcal{E}(x_{high})]
    \label{eqa:concat}
\end{equation}
where $c_{f}$ represents the condition for diffusion model, $\mathcal{E}$ denotes the VAE encoder. To leverage the external priors of denoising UNet, we adopt a method similar to \cite{b21,b29} by introducing an adaptor $\mathcal{N}$ that creates a trainable copy of the UNet. This adaptor $\mathcal{N}$ enables the injection of the conditioning prompt $c_{f}$ into the diffusion process as structural guidance.

Finally, we finetune the diffusion model via optimizing following objective function:
\begin{equation}
\mathcal{L}_{SPR} = \mathbb{E}_{x_{in}, t, c_f, \epsilon \sim N(0, 1)} \left[ \|\epsilon - \epsilon_{\theta}(z_t, t, \mathcal{N}(c_f))\|_2^2 \right]
\end{equation}
where $\epsilon_{\theta}$ denotes the pretrained denosing UNet. During finetuning, we only apply gradient descent to the learnable kernel $\text{conv}_{1\times 1}$ and the adaptor $\mathcal{N}$ while freezing the weights of $\epsilon_{\theta}$.

\subsection{Haze-aware Self-correcting Refiner (HCR)} \label{HCR}
Despite the improved faithfulness of our SPR-boosted dehazing process in the latent space, the final dehazed image often diverges from the groundtruth, \textit{i.e., hallucination} dilemma. Previous works \cite{b5,b32} address the challenges by finetuning the decoder $\mathcal{D}$ with a small refine network. Building on this approach, we propose a self-correction mechanism that leverages haze-aware priors within a hazy image to enhance fidelity. This is achieved by inserting several self-attention (SA) blocks into $\mathcal{D}$, modulated by priors that recognize haze density (see Fig.\ref{overview}). These priors are selective in prioritizing areas with thinner haze while deprioritizing areas with dense haze. 
%These priors, capable of recognizing haze density, are injected into $\mathcal{D}$ and adaptively modulate its attention maps. They are selective in prioritizing areas with thinner haze while deprioritizing areas with dense haze. 
Thus, self-correction is enabled by distribution alignment towards more reliable signals, such as clearer regions.

%To address these challenges, we propose a Haze-aware Prompted Self-correction Refine Network (HCRN), inspired by \cite{b5,b32}, designed to refine VAE decoder features by incorporating self-correcting priors derived from VAE encoder features. In this process, we introduce haze-aware internal priors to guide and enhance the refinement. Unlike prior works \cite{b5,b32}, which perform straightforward feature calibration, our approach selectively aligns encoder and decoder features to ensure more precise and effective refinement. By deprioritizing features from haze-dominant regions, our method minimizes their influence on the final reconstruction, leading to cleaner and more faithful dehazing results.

To initialize the haze-aware internal prior, we employ the Dark Channel Prior (DCP) \cite{b33}, which provides an efficient yet approximate estimation of haze distribution within an image. From the input image \( x_{in} \), we compute the DCP mask \( M_{DCP} \in \mathbb{R}^{H \times W \times 1} \), where \( (H, W) \) represents the spatial dimensions of \( x_{in} \). Higher values in \( M_{DCP} \) indicate regions with denser haze. The mask is then flattened into a 1D vector \( \overline{M}_{DCP} \in \mathbb{R}^{N \times 1} \), where \( N = H \times W \).

In the \( l \)-th block of the decoder \( \mathcal{D} \), we compute a correlation map \( M_{corr,l} \) to capture interactions among all elements in \( \overline{M}_{DCP} \). This is achieved by applying the cross product between \( \overline{M}_{DCP} \) and its transposition, modulated through two learnable weight matrices \( W^{Q}_l \in \mathbb{R}^{1 \times N_{l}} \) and \( W^{K}_l \in \mathbb{R}^{1 \times N_{l}} \):
% \textcolor{red}{(specify the equation according to your implementation!)}:
\begin{equation}
M_{corr,l} = (\overline{M}_{DCP} W^Q_l) \times (\overline{M}_{DCP} W^K_l)^T
\end{equation}
where $N_{l} = H_{l}W_{l}$  denotes the size of the feature map in the $l$-th block. Then a top-$k$ selection \cite{b56}, denoted as $\text{Top}_k(\cdot)$, is applied to $M_{corr,l}$ to extract the indices of the $k$ highest values in it: 

\begin{equation}
\mathcal{I} = \{ (i, j) \mid M_{corr,l}^{ij} \in \text{Top}_k({M_{corr,l}}) \}
\label{eqa:topk}
\end{equation}
% \textcolor{red}{To be modified}\\
%the indices $index_h$ of the $k$ most correlated regions. As $M_{corr}$ captures the interactions between different regions of the DCP mask, a high value in $M_{corr}$ indicates strong correlations between two regions in terms of haze density. Consequently, the indices $index_h$ correspond to haze-dominant areas, identifying the regions most affected by haze.
%\textcolor{red}{(why?, to be clarified, you need to firstly mention what high value means in DCP mask in the previous paragraph, and then explain what high value means in correlation matrix)}.

We use the set of indices \(\mathcal{I}\), corresponding to the \(k\) most haze-affected regions, to sparsify \(M_{corr}\), yielding a sparsed mask \(M_s\). The elements of \(M_s\) are computed as follows:

\begin{equation}
M_s^{ij} =
\begin{cases} 
-\infty, & (i,j) \in \mathcal{I} \\[8pt]
1 - M_{corr}^{ij}, & (i,j) \notin \mathcal{I}.
\end{cases}
\end{equation}

%This formulation ensures that regions in \(\mathcal{I}\) (haze-dominant regions) are excluded from subsequent computations by assigning them a value of \(-\infty\), while less haze-affected regions are weighted based on their correlation values.
%\textcolor{red}{(rewrite the above quation by new math symbols, specifiy what s r, how to choose r.)} 

This sparsification strategy selectively eliminates the negative impact of haze-dominant areas in $M_{corr}$ %significantly downweighing 
by assigning \(-\infty\), while emphasizing the representation of clearer areas by assigning a weight of $1 - M_{corr}^{ij}$. Next, $M_{s}$ is injected to modulate the standard self-attention calculation in the Window Swin Transformer (WST) block in the decoder $\mathcal{D}$ by:
%The window-based self-attention (WSA) is subsequently computed as:

\begin{equation}
attn(M_{s}) = \text{softmax}((Q K^T) \odot M_s / \sqrt{N_l}) V
\label{eqa:att}
\end{equation}
where \(\odot\) denotes element-wise multiplication, and \(Q\), \(K\), and \(V\) are the Query, Key, and Value matrices of attention layer. Note that we adopt window Swin Transformer blocks \cite{b42} of size $(H_{win}, W_{win})$ for computation efficiency. Consequently, the dimensions of $M_{DCP}$ are adapted to match this window size. The transformer blocks are inserted into the two lowest-resolution stages of $\mathcal{D}$. 
%Meanwhile, we apply the same window partitioning strategy to $M_{DCP}$ to ensure alignment with the WST for seamless integration.
%This attention modulation trains the network to focus on more reliable signals —specifically, clear regions. 
%In this stage, we jointly train the refinement network and fine-tune the decoder \(\mathcal{D}\), we first generate training data by sampling from the finetuned SPR, to produce the latent space vector \(z_0\). Additionally, we extract features \(F_{\mathcal{E}}(x_{in})\) from the input \(x_{in}\) using the VAE encoder \(\mathcal{E}\). During this training phase, \(z_0\) is decoded by \(\mathcal{D}\) to reconstruct the output \(x_r\), while the decoded features \(F_{\mathcal{D}}\) are refined at each resolution level by aligning them with the corresponding features \(F_{\mathcal{E}}(x_{in})\) in the refine network, following the methodology outlined in \cite{b32}. The training process minimizes the following objective function:
During training, we generate the latent vector \(z_0\) from the finetuned SPR. Additionally, we extract multi-scale features \(F_{\mathcal{E}}(x_{in})\) of the input \(x_{in}\) from the VAE encoder \(\mathcal{E}\). The output dehazed image \(x_r\) is:

\begin{equation}
x_r = \mathcal{D}\left(z_0; R(F_{\mathcal{E}}(x_{in}), F_{D}), attn(M_{s}) \right)
\end{equation}
where $R$ represents the trainable refine network that aligns $F_{\mathcal{E}}(x_{in})$ and the decoder feature $F_D$, generating the refined feature $F_{RN}$. To train HCR, we minimize the following loss:

\begin{equation}
\mathcal{L}_{HCR} = ||x_{r}-x_{GT}||_{1} + \mathcal{L}_{VGG}(x_{r},x_{GT}) + \mathcal{L}_{\text{adv}}(x_r,x_{GT})
\end{equation}
where $x_{GT}$ denote the ground true image, $\mathcal{L}_{VGG}$ and $\mathcal{L}_{\text{adv}}$ are the VGG perceptual loss \cite{b79} and the adversarial loss \cite{b81}.

\begin{table*}[h!]
\caption{Quantitative comparison on five real-world datasets, \textbf{bold} and \underline{underline} indicate the best and the second-best, respectively}
\vspace{-0.2cm}
\centering
\label{quant1}
\begin{tabular}{c|c|c c c c c c c c c c c}  % Added two extra 'c' for the new columns
\hline
Dataset & Metrics & FCB & RIDCP & C2PNet & DEANet & FSNet & SFSNiD& DiffUIR & CASM &PromptIR &FPro&Ours \\ \hline
\multirow{5}{*}{I-Haze}   & PSNR$\uparrow$ & 17.35 & \underline{17.41}  & 16.70 & 16.54 & 17.13 & 16.18& 16.75 & 17.12 &16.07&16.69& \textbf{21.69} \\ 
   & SSIM$\uparrow$  & \underline{0.82}  & 0.79 & 0.77 & 0.78 & 0.78 & 0.74 & 0.78 & 0.80&0.76&0.77&  \textbf{0.87} \\ 
   & CIEDE$\downarrow$  &\underline{11.53} & 12.14 & 12.52 & 13.05 & 12.17 & 15.64 & 12.36&12.62& 13.81&13.05&\textbf{7.88}  \\ 
   & CLIPIQA$\uparrow$    & 0.31  & 0.35& 0.34 & 0.35 & 0.35 & 0.24 & 0.34
    &\underline{0.36}&0.34&0.24&\textbf{0.40} \\
   & NIMA$\uparrow$  & 5.01   & \underline{5.12} & 4.45 & 4.55 & 4.38 & 4.00 & 4.43&5.10&4.72&4.77&\textbf{5.27} \\ \hline
\multirow{5}{*}{O-Haze}   & PSNR$\uparrow$ & 17.39 & 16.74  & 16.53 & 16.53 & 16.01 & \underline{17.75} &15.33 &17.64&16.41&17.08& \textbf{18.73} \\ 
   & SSIM$\uparrow$   & \textbf{0.80} &  0.76 & 0.70 & 0.69 & 0.67 & 0.66 &0.65 & \underline{0.77}&0.68&0.71&\underline{0.77} \\ 
   & CIEDE$\downarrow$  & 13.58 & \underline{12.03} & 15.98 & 15.89 & 15.84 & 14.32 &16.69&12.98&16.01&15.52&\textbf{11.65} \\ 
   & CLIPIQA$\uparrow$ & 0.41 & \textbf{0.52} & \underline{0.44} & 0.43 & \underline{0.44} & 0.30&0.42 &\underline{0.44}&0.46&0.33& \underline{0.44} \\
   & NIMA$\uparrow$  & \underline{5.12} & \textbf{5.15} & 4.97 & 4.96 & 4.95 & 4.38 &4.85&5.13 & 5.12&5.06&\textbf{5.15} \\ \hline
\multirow{5}{*}{DenseHaze}   & PSNR$\uparrow$   & 13.16 & 11.28 & 11.67 & 11.57 & 11.34 & 12.35 &10.23&\underline{13.98} & 11.54&11.12& \textbf{14.06} \\ 
   & SSIM$\uparrow$   & 0.49 & \underline{0.49} & 0.44  & 0.44 & 0.43 & 0.43 &0.40&\underline{0.50}&0.43&0.46&\textbf{0.52} \\ 
   & CIEDE$\downarrow$  & 21.32 & 25.03 & 24.51 & 24.84 & 24.87 & 22.82 &26.96&\underline{18.58}&24.84&25.14&\textbf{18.53} \\ 
   & CLIPIQA$\uparrow$    & 0.21 & \textbf{0.33} & 0.25 & 0.23 & 0.25 & 0.23  &0.25&0.19&0.24&0.24&\underline{0.31} \\
   & NIMA$\uparrow$  & 5.00   & \underline{5.24}  & 4.82 & 4.84 & 4.80  &4.08 &4.64&4.09&4.86&4.63&\textbf{5.25} \\ \hline
\multirow{5}{*}{NhHaze}   & PSNR$\uparrow$   & \underline{14.17}  & 12.95 & 12.47  & 12.37 & 12.27 & 13.20 &11.97&13.61&12.30&12.40&\textbf{14.62} \\ 
   & SSIM$\uparrow$   & \textbf{0.62} & 0.48 & 0.53 & 0.52 & 0.51 & 0.49  &0.49 &\textbf{0.62}&0.51&0.51&\underline{0.58} \\ 
   & CIEDE$\downarrow$  & \underline{17.30} & 19.91 & 21.86 & 18.59 & 21.48 & 20.17&21.63&17.98&21.77&21.66&\textbf{16.73} \\ 
   & CLIPIQA$\uparrow$    & 0.39  & \textbf{0.49} & 0.42 & 0.40 & \underline{0.43} & 0.35&0.40&0.42&\textbf{0.49}&0.37&\underline{0.43} \\
   & NIMA$\uparrow$  & 5.56 & \textbf{5.76} & 5.65 & 5.60 & 5.64 & 4.94  &5.56 &5.08&\underline{5.69}&5.62&5.61 \\ \hline
\multirow{2}{*}{RTTS}   & CLIPIQA$\uparrow$ & 0.23 & \textbf{0.34}  &  0.32 & 0.31 &  \underline{0.33} & 0.28&\underline{0.33} & 0.30&\underline{0.33}&0.32&\textbf{0.34} \\ 
  
   & NIMA$\uparrow$  & 4.52 &  \underline{5.13} & 5.03 & 5.02 & 4.97 & 4.67& 5.02&5.02  & 4.96&5.01&\textbf{5.14}\\ \hline
\end{tabular}
\vspace{0.1cm}

\vspace{-0.7cm}
\end{table*}

% \begin{table*}[h!]
% \label{quant2}
% \centering
% \begin{tabular}{c|c|c c c c c c c c c c c}  % Added two extra 'c' for the new columns
% \hline
% Dataset & Metrics & FCB & RIDCP & C2PNet & DEANet & FSNet & SFSNiD & DiffUIR & CASM &PromptIR &FPro& Ours \\ \hline
% \multirow{2}{*}{RTTS}   & CLIPIQA$\uparrow$ & 0.23 & \textbf{0.34}  &  0.32 & 0.31 &  \underline{0.33} & 0.28&\underline{0.33} & 0.30&\underline{0.33}&0.32&\textbf{0.34} \\ 
  
%    & NIMA$\uparrow$  & 4.52 &  \underline{5.13} & 5.03 & 5.02 & 4.97 & 4.67& 5.02&5.02  & 4.96&5.01&\textbf{5.14}\\ 

% \hline

% \end{tabular}
% \vspace{0.1cm}
% \caption{Quantitative comparison on Real-world RTTS Dehazing}
% \end{table*}

%\subsection{Training Strategy}

\section{Experiments}
\vspace{-0.1cm}
%\textcolor{red}{(Optionally removed)} 
% In this section, we present a comprehensive evaluation of the proposed model's performance on real-world dehazing tasks. Section IV.A outlines the experimental settings, followed by a comparative analysis in Section IV.B, where the proposed model is benchmarked against multiple state-of-the-art dehazing methods across various datasets. Finally, Section IV.C provides detailed ablation studies to assess the effectiveness of the proposed modules.
\subsection{Experimental Settings}
\noindent\textbf{Datasets.} For the training set, we utilize both the Indoor Training Set (ITS) and Outdoor Training Set (OTS) from the REalistic Single Image DEhazing (RESIDE) \cite{b43} benchmark. For the test sets, we select four commonly used datasets of paired hazy-clean images: I-Haze \cite{b47}, O-Haze \cite{b44}, Dense-Haze \cite{b45}, and NH-Haze \cite{b46}. Additionally, we include the RTTS \cite{b43} dataset, which contains only real-world hazy images without groundtruth, to evaluate the model's dehazing performance for real scenes.

We utilize the ASM-based synthesis pipeline \cite{b7} to generate the synthetic pairs for training.
%\textcolor{red}{(add citation)}. 
To demonstrate the generalization capability of our model, we train exclusively on synthetic hazy-clean pairs from ITS and OTS, and evaluate directly on the test sets without any ad-hoc finetuning. %During training, we utilize the widely adopted ASM-based synthesis pipeline to generate the synthetic paired training dataset.

\noindent\textbf{Evaluation.} We adopt widely used reference-based metrics---PSNR, SSIM, and CIEDE \cite{b4}---to evaluate image distortion, structural similarity, and color
difference, respectively, between the dehazed and groundtruth images.
To further assess the perceptual quality of the dehazed images, we employ CLIPIQA \cite{b5} and NIMA \cite{b6}, which are non-reference image quality assessment metrics based on pretrained models.

\noindent\textbf{Benchmark Comparison.}
We evaluate the performance of our proposed model against a range of state-of-the-art dehazing algorithms, focusing on diffusion-based methods (e.g. FCB \cite{b7}, DiffUIR \cite{b66}), real-scene dehazing methods (e.g. RIDCP \cite{b12}, CASM \cite{b8}, SFSNiD \cite{b11}), image prompting (e.g. PromptIR \cite{b2}, FPro \cite{b3}), as well as other state-of-the-art algorithms (e.g. C2PNet \cite{b55}, DEANet \cite{b10}, FSNet \cite{b9}). We specify our implementation strategy in Supplementary Materials I.
%\textcolor{red}{(Too many categories! Reduce! emphasize on diffusion methods, real-scene methods and prompting strategies. What is the point to compare with frequency-, prior- and unified- methods?)} Additionally, we compare with approaches tailored for real-world dehazing, such as RIDCP \cite{b12}, which requires ad-hoc training on real-world datasets. 
%\textcolor{red}{(consider to move this paragraph to Appendix) 

\vspace{-0.1cm}
\subsection{Quantitative Comparison}
\vspace{-0.1cm}
\begin{figure*}[]
    \centering
    \includegraphics[width=\textwidth]{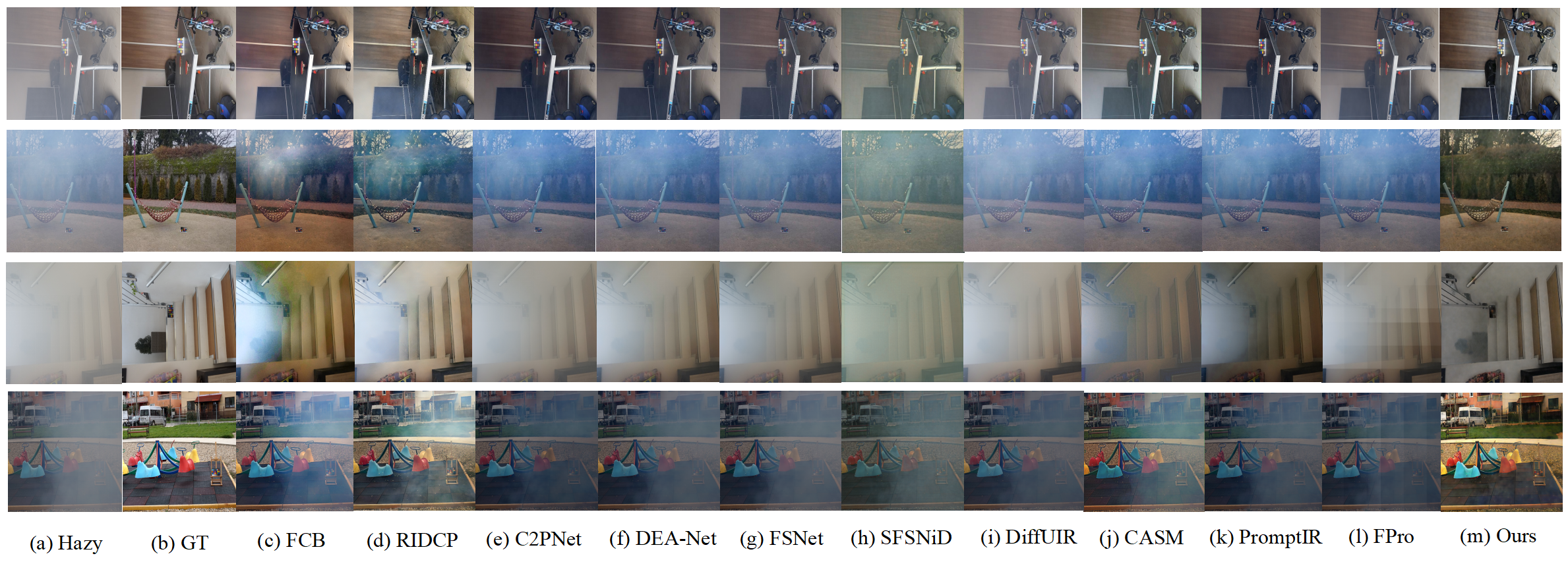} 
    \vspace{-0.6cm}
    \caption{Qualitative comparison of dehazing results with different methods. From top to bottom are the results on I-Haze, O-Haze, DenseHaze and NhHaze.}
    \vspace{-0.3cm}
    \label{fig:qual}
\end{figure*}

As presented in Table I, our proposed method consistently outperforms all baseline methods across the test datasets. Notably, referenced
metrics such as PSNR, SSIM and CIEDE excel other methods significantly. This can be attributed to the effective integration of internal priors alongside external priors from the pre-trained LDM, which significantly enhances fidelity while preserving perceptual quality. As for real-world dehazing, our method demonstrates superiority on RTTS dataset, according to the improvements in the non-reference NIMA and CLIPQA metrics. Notably, even without finetuning on real-world hazy images, our approach achieves competitive results to RIDCP \cite{b12}, which is tailored for such scenarios. 

\begin{table}[t]
\caption{Ablation study of SPR and HCR on I-Haze and O-Haze.}
\setlength{\tabcolsep}{4.5pt}
\centering
\fontsize{5pt}{6pt}\selectfont
\begin{tabular}{c|c c|c c c|c c c} 
\hline
\multirow{4}{*}{\textbf{Exp.}} & \multicolumn{5}{c|}{\textbf{Modules}} & \multicolumn{3}{c}{\textbf{I-Haze / O-Haze}} \\ \cline{2-5} \cline{6-9}
 & \makecell[c]{Vanilla\\Adapter} & \makecell[c]{Adaptor \\w/ SPR} & \makecell[c]{Vanilla\\Refine\\Network} & \makecell[c]{HCR \\w/o $M_s$} & \makecell[c]{HCR \\w/$M_s$} & \textbf{PSNR$\uparrow$} & \textbf{SSIM$\uparrow$} & \textbf{CIEDE$\downarrow$} \\ \hline
 
 (a)& \checkmark &   & & & &17.36 / 16.43 &  0.74 / 0.46 & 13.29 / 16.22 \\
 (b)&  & \checkmark & & & &19.33 / 16.97 & \underline{0.86} / 0.48 & 10.02 / 14.96 \\ \hline
 (c)& \checkmark & & \checkmark & & &19.90 / 18.35 & 0.75 / \underline{0.75} & \underline{9.45} / 12.56 \\ 
 (d)&  & \checkmark & \checkmark & & &19.94 / 18.55 & 0.84 / \textbf{0.77} & 10.47 / 12.54 \\
 (e)&  & \checkmark & & \checkmark & &\underline{20.15} / \underline{18.56} & 0.84 / \textbf{0.77} & 10.25 / \underline{12.44} \\
 (f)&&\checkmark&&&\checkmark&\textbf{21.69} / \textbf{18.73} & \textbf{0.87} / \textbf{0.77} & \textbf{7.88} / \textbf{11.65}\\ \hline
 
\end{tabular}
\vspace{-0.2cm}
\label{table:abl}
\end{table}

\vspace{-0.1cm}
\subsection{Qualitative Comparison}
\vspace{-0.1cm}
Fig.~\ref{fig:qual} presents a visual comparison of dehazing results against existing benchmark methods. On the I-Haze and O-Haze datasets, our method shows superior performance, delivering visually compelling results with significantly reduced haze residue. For the DenseHaze and NhHaze datasets, our approach effectively removes the majority of the haze and accurately restores the underlying structure of the input images. Notably, the proposed HCR module has substantially improved color fidelity, effectively correcting color bias and suppressing the hallucinated details from the pretrained LDMs. %Consequently, our method achieves more accurate and visually compelling reconstruction outcomes compared to end-to-end approaches, while introducing fewer unwanted artifacts than other external prior-based methods. 
We provide visual results on RTTS in Supplementary Materials III.
\vspace{-0.1cm}
\subsection{Ablation Study}
\vspace{-0.1cm}
\label{abl:spr}
% \textcolor{red}{Merge Table 3 and Table 4 to save some space!}

% \textbf{Overview}\\

\begin{figure}[t]
    \centerline{\includegraphics[height=1.75cm]{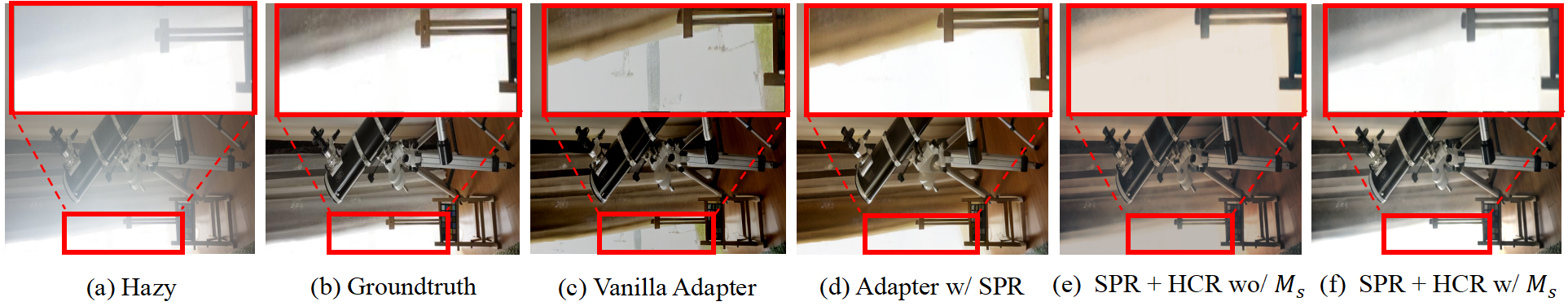}}
    \caption{Ablation study on the proposed method. (c) finetuning only the vanilla adapter $\mathcal{N}$. (d) finetuning SPR only. (e) finetuning both SPR and HCR, but without $M_s$ modulation (f) full setting.}
    \label{fig:abl1}
\end{figure}

\noindent\textbf{Effect of SPR.} 
To show the importance of the proposed SPR, we present quantitative results in Table III and visual results in Fig.\ref{fig:abl1}. As shown in Table III (a) and (b), SPR significantly enhances performance across test sets compared to a vanilla adapter without structural prompts. 
The visual results in Fig. \ref{fig:abl1}(c) contain hallucinated details and incorrect content, while (d) exhibits a faithful restoration. 
%The visual result in Fig.\ref{fig:abl1}(d) exhibit fewer hallucinated details and unrealistic artifacts compared to (c), leading to more faithful restoration.
% \textcolor{red}{(consider to remove this sentence if no space left.) 
Furthermore, as shown in Table III (c) and (d), such improvement persist after integrating a refine network. Overall, these findings validate the role of SPR in restoring faithful structural information via the proposed prompt.
They demonstrate that high fidelity can be achieved in a well-prompted latent space, even without decoder refinement.

\noindent\textbf{Effect of HCR.} Table III (e) and (f) show that the HCR module markedly improves PSNR, underscoring its strength in enhancing faithfulness. The improvement in CIEDE further highlights HCR's ability to mitigate color bias. Additionally, Table III (d) and (e) show that such improvement arises from $M_s$ modulation via haze-aware prior, while merely inserting WST blocks yields only minimal impact on the results.
%while Table III (d) and (e) show that the injected WST has minimal impact on the results, with the $M_s$ modulation in the HCR playing the key role. Additionally, the significant improvement in CIEDE highlights the effectiveness of HCR in mitigating color bias. 

To further validate the effectiveness of haze-aware prior on color correction, we provide visual comparisons in Fig.~\ref{fig:abl1}. Without the haze-aware prior, noticeable color shifts are evident in Fig.~\ref{fig:abl1}(e). In contrast, the color bias is substantially reduced with the integration of haze-aware prior in Fig.~\ref{fig:abl1}(f).

\section{Conclusion}
\vspace{-0.1cm}
We propose a novel real-world dehazing framework ProDehaze that directs external image priors within pretrained diffusion models with internal priors for faithful dehazing. Specifically, we design two selective priors to prompt the model: structural priors in the latent space and haze-aware prompts during decoding. These priors enable the pretrained model to produce high-fidelity results with preserved structures, faithful details, and reduced color bias. Our framework demonstrates the effectiveness of guiding pretrained knowledge toward critical image areas for faithful restoration, offering a promising avenue for addressing other inverse problems.

\footnotesize

\bibliographystyle{IEEEtran}
\bibliography{main.bib}

\end{document}